\documentclass[sigconf]{acmart}
\AtBeginDocument{%
  \providecommand\BibTeX{{%
    \normalfont B\kern-0.5em{\scshape i\kern-0.25em b}\kern-0.8em\TeX}}}

\copyrightyear{2020}
\acmYear{2020}
\setcopyright{acmcopyright}\acmConference[MM '20]{Proceedings of the 28th ACM
International Conference on Multimedia}{October 12--16, 2020}{Seattle, WA, USA}
\acmBooktitle{Proceedings of the 28th ACM International Conference on Multimedia
(MM '20), October 12--16, 2020, Seattle, WA, USA}
\acmPrice{15.00}
\acmDOI{10.1145/3394171.3413592}
\acmISBN{978-1-4503-7988-5/20/10}

\usepackage{dsfont}
\begin{document}
\fancyhead{}
\title{Exploring Font-independent Features for Scene Text Recognition}


\author{Yizhi Wang}
\affiliation{\institution{Wangxuan Institute of Computer Technology, Peking University}
\country{China}}
\email{wangyizhi@pku.edu.cn}

\author{Zhouhui Lian}
\authornote{Corresponding author}
\affiliation{\institution{Wangxuan Institute of Computer Technology, Peking University}
\country{China}}
\email{lianzhouhui@pku.edu.cn}

\renewcommand{\shortauthors}{Wang and Lian}

\begin{abstract}
Scene text recognition (STR) has been extensively studied in last few years.
Many recently-proposed methods are specially designed to accommodate the arbitrary shape, layout and orientation of scene texts, but ignoring that various font (or writing) styles also pose severe challenges to STR.
These methods, where font features and content features of characters are tangled, perform poorly in text recognition on scene images with texts in novel font styles.
To address this problem, we explore font-independent features of scene texts via attentional generation of glyphs in a large number of font styles.
Specifically, we introduce trainable font embeddings to shape the font styles of generated glyphs, with the image feature of scene text only representing its essential patterns.
The generation process is directed by the spatial attention mechanism, which effectively copes with irregular texts and generates higher-quality glyphs than existing image-to-image translation methods.
Experiments conducted on several STR benchmarks demonstrate the superiority of our method compared to the state of the art.
\end{abstract}



\maketitle

\section{Introduction}

\begin{figure}[t!]
  \centering
  \includegraphics[width=\columnwidth]{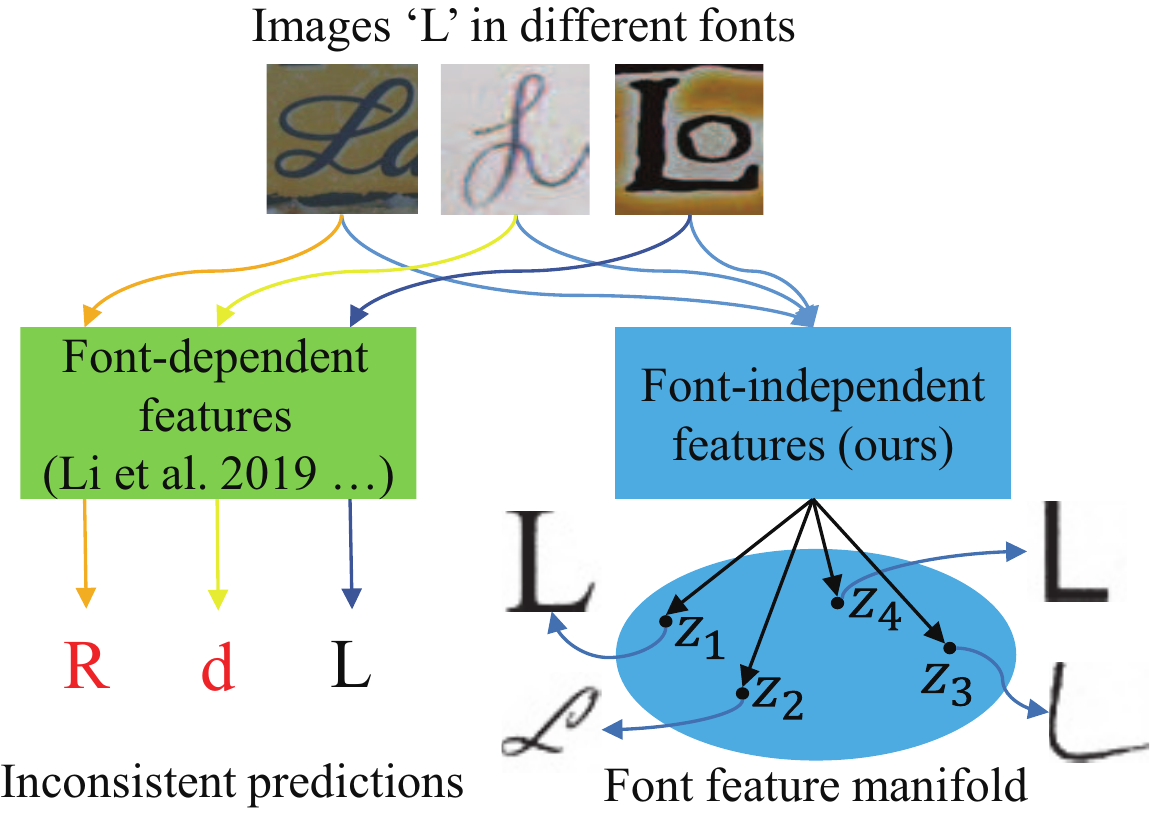}
  \caption{Exploring font-independent features makes our model generalize better on text images whose font styles are rarely or never seen in the training dataset.
  In contrast, most existing methods, such as~\cite{li2019show,shi2018aster}, tend to make wrong predictions on text images in novel font styles.
  We introduce trainable font embeedings serving as the font features of generated glyphs (shown in the bottom right), so as to decrease the unnecessary font style information in scene text features.
  }
  \label{fig:Introduction}
\end{figure}

Feature representation plays a crucial role in scene text recognition.
Before the popularity of deep learning, most methods employ HOG (Histogram of Oriented Gradient)~\cite{dalal2005histograms} like features for text recognition, but these handcrafted features are not able to satisfactorily deal with noisy data.
Afterwards, deep learning techniques have been widely adopted for scene text recognition achieving impressive performance~\cite{jaderberg2015deep,jaderberg2016reading}.
At present, most prevalent models are devised by combining convolutional neural networks (CNNs) and recurrent neural networks (RNNs), such as~\cite{cheng2017focusing,shi2018aster,li2019show,yang2019symmetry}.
The CNN extracts visual cues within images and the RNN generates character sequences on the basis of CNN features.
There are also some works~\cite{fang2018attention,sheng2018nrtr} proposing to replace RNNs with fully convolutional networks or self-attention mechanism~\cite{vaswani2017attention}.
In the training phase, these models usually employ the softmax classifier and cross-entropy loss function, following the common object recognition or classification frameworks.
However, there is no explicit mechanism in these frameworks to guarantee the removal of font style information from the learnt features.
In other words, these frameworks have no clear idea of what the font style and content of a text image are, which weakens their generalization ability.
As a result, these models tend to fail in extracting discriminative features for accurately recognizing text images in novel font styles, as is shown in Figure~\ref{fig:Introduction}.
A recent survey paper~\cite{baek2019wrong} supports our opinions by reporting that “difficult fonts” remains a challenging and ongoing problem in STR.
\par
In this paper, we investigate how to extract font-independent features from scene texts, so that our model generalizes well on those scene texts in novel fonts.
Specifically, we introduce trainable font embeddings, which are concatenated with the scene text features, to generate glyphs in various font styles.
The font embeedings are trained to serve as the font features of target glyphs, so as to decrease the unnecessary font style information in scene text features.
To accommodate the arbitrary shape and layout of scene texts, we follow the guidance of spatial attention mechanism to generate glyphs one by one.
In each generation step, we employ a GAN (Generative Adversarial Network)~\cite{goodfellow2014generative} based generator to translate CNN features from a focusing position into glyphs of multiple fonts.
Through our experiments, we find that our technique makes the model less sensitive to font variance, and markedly enhances the STR performance.~\footnote{Source code is available at \url{https://actasidiot.github.io/EFIFSTR/}}

\begin{figure*}[t!]
  \centering
  \includegraphics[width=16cm]{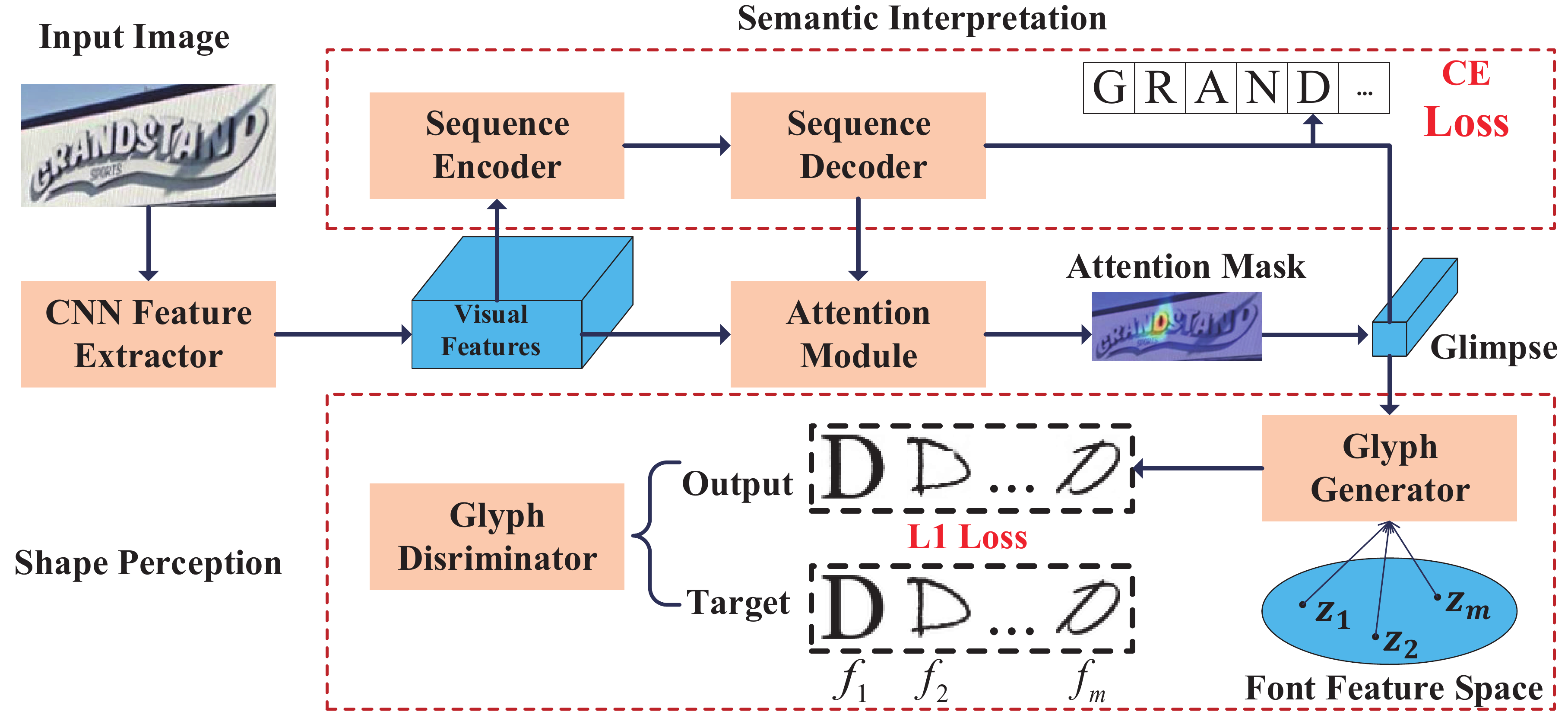}
  \caption{The pipeline of our proposed method.}
  \label{fig:Pipeline}
\end{figure*}

\section{Related Work}
\subsection{Regular Scene Text Recognition}
Considering there is a considerable body of literature on STR, we only discuss works that are most closely related to ours.
\cite{wang2012end,jaderberg2015deep,jaderberg2016reading} are among the early works in using deep convolutional neural networks as feature extractors for STR.
\cite{he2016reading} and \cite{shi2017an} considered words as one-dimensional sequences of varying lengths, and employed RNNs to model the sequences without explicit character separation.
A Connectionist Temporal Classification (CTC) layer was adopted to decode the sequences.
Inspired by the sequence-to-sequence framework for machine translation, \cite{lee2016recursive} and \cite{shi2016robust} proposed to recognize text using an attention-based encoder-decoder framework.
\subsection{Irregular Scene Text Recognition}
\cite{shi2016robust,shi2018aster} proposed an explicit rectification mechanism, which is based on Thin Plate Spline, to rectify distorted and curved texts for recognition.
However, many texts in the wild have arbitrary shapes and layouts, which makes it difficult to transform them into horizontal texts through their proposed interpolation methods.
\cite{liu2018char} presented the Char-Net to detect and rectify individual characters, which, however, requires extra character-level annotations.
\cite{cheng2018aon} applied LSTMs~\cite{hochreiter1997long} in four directions to encode arbitrarily-oriented text.
\cite{li2019show} adopted a 2D attention based encoder-decoder network for irregular text recognition inspired by~\cite{xu2015show}.
\cite{sheng2018nrtr} introduced the Transformer~\cite{vaswani2017attention} model into STR which can be trained in parallel and good at capturing dependency relationships in sequence.
\par
\subsection{Generative Models for Scene Text Recognition}
Generative models, such as the Bayesian Network and Generative Adversarial Network (GAN), model the distribution of individual classes while discriminative models learn the (hard or soft) boundary between classes.
There is a growing trend that generative models are introduced into scene text recognition in many recent works, such as~\cite{george2017generative,liu2018synthetically,wang2019boosting,zhan2019spatial,zhan2019ga}.
Previous to our work, ~\cite{liu2018synthetically,wang2019boosting} proposed to extract more robust features by mapping scene texts into canonical glyphs.
~\cite{liu2018synthetically} proposed to transform the whole scene text image into corresponding horizontally-written canonical glyphs for promoting feature learning.
Through their experiments, the guidance of canonical forms of glyphs is proved to be effective for feature learning in STR.
\cite{wang2019boosting} utilized glyphs in four fonts as targets to generate, employing random vectors as the font embeddings.
However, these font embeddings were fixed in the training phase and could not appropriately serve as the font features of target glyphs.
Without reliable font features as embeddings, the feature extractor of scene texts could be distracted from extracting font-independent features.
Besides, both~\cite{liu2018synthetically} and~\cite{wang2019boosting} employed the basic CNN-DCNN (DeConvolutional Neural Network) framework which cannot cope well with irregular texts.
Motivated by these analyses, we propose a novel method, i.e., attentional glyph generation with trainable font embeddings, to overcome the deficiencies of above-mentioned methods.
\par
\subsection{Multi-font Character Recognition}
It is worth noting that most character (text) recognition methods, which are trained with (a sufficient number of) multiple font images, have the same purpose to extract the font-independent feature.
\cite{zhong2015multi} proposed the multi-pooling operation for CNN to increase its robustness to some simple font transformation.
However, the learning of font-independent features is still very difficult for existing methods, with the character category labels as the only guidance.
In Figure 1, the letter `L’ in different font styles causes many existing models to make wrong predictions.
A simple idea is providing the font labels of the training images for CNNs to learn. However, the annotation cost will be huge, especially for real-world images.
Instead, we instruct CNNs to learn the most “essential” features for reconstructing glyphs in various font styles.
We utilize glyphs in ~\textbf{a large number of} fonts as explicit guidance (compared to the training images, the number of these glyph images can be ignored) and achieve significant recognition improvement.

\section{Method Description}
\subsection{Overview}
We first briefly illustrate the pipeline of proposed model shown in Figure~\ref{fig:Pipeline}.
Given an input image $x \in \mathbb{R}^{H_{0} \times W_{0} \times C_{0}}$, we first employ a CNN Feature Extractor to extract its visual feature $F(x) \in \mathbb{R}^{H \times W \times C}$.
The above-mentioned three dimensions represent height, width and channel number, respectively
($H > 1$ and $W> 1$ for preserving more spatial information of the input image).\par
Then we send the CNN features $F(x)$ into the Sequence Encoder and Decoder (such as LSTM~\cite{hochreiter1997long} and Transformer~\cite{vaswani2017attention}) for sequence modeling.
During the decoding step $t$, the CNN feature maps $F(x)$ and the hidden layer's output of Sequence Decoder $h(x, t)$ are altogether fed into the attention module to calculate the attention mask $M(x, t) \in \mathbb{R}^{H \times W}$:
\begin{equation}
M(x, t) = Attention(F(x), h(x, t)),
\end{equation}
deciding which position is supposed to be paid more attention to at this moment.
Afterwards, by multiplying visual features $F(x)$ by the attention mask over all channels, we get a weighted vector:
\begin{equation}
c(x,t) = M(x, t) \cdot F(x),
\end{equation}
which is commonly known as ``the glimpse vector''.
\par
The next process, where our main contribution lies, is utilizing the glimpse vector to generate its multi-font glyphs and predict character symbols synchronously.
A Glyph Generator based on GAN is adopted for generating glyphs:
\begin{equation}
\label{equ:GlyGen}
\hat{g}_{k}(x,t) = GlyphGen([c(x,t); z_{k}]), 1 \leq k \leq m,
\end{equation}
where $z_{k} \in \mathbb{R}^{C^{'}}$ denotes font embedding, which determines the font style of target glyphs; the square bracket denotes concatenation; $m$ is the number of selected target fonts.
Different from Conditional GAN~\cite{mirza2014conditional}, $z_{1},z_{2},...z_{m}$ will be treated as trainable variables to better serve as the font conditions of glyphs.
Meanwhile, $h(x, t)$ and $c(x,t)$ are taken for predicting the current-step symbol:
\begin{equation}
p(y_{t}) = softmax(W_{o}[h(x, t);c(x,t)] + b_{o}).
\end{equation}
The two learning branches work cooperatively and synchronously, contributing to the enhancement of scene text feature learning.
\subsection{CNN Feature Extractor}
\label{Sec:CNNModel}
The CNN Feature Extractor of our model is adapted from~\cite{shi2018aster}.
The difference is that the vertical dimension of CNN feature maps will not be down-sampled to 1, by keeping the stride $1 \times 1$ in the fourth and fifth residual blocks.
Then the output feature $F(x)$ has the shape of $H \times W \times C$ where $H = H_{0} / 8$ and  $W = W_{0} / 4$.
\subsection{Sequence Encoder-Decoder and Attention Module}
The attention module intends to output an attention mask according to the hidden layer of LSTM or Transformer and encoded visual features.
We try several popular methodologies for calculating the attention mask, including the 2D-Attention mechanism proposed in~\cite{li2019show}, the traditional 1D-attention employed in~\cite{shi2018aster} and the self-attention mechanism employed in~\cite{sheng2018nrtr}.
Based on the experimental results (will be shown in the Section~\ref{Sec:AblaStudy}), we adopt the scheme in~\cite{li2019show} which results in the best performance for our framework.\par
Here we briefly review the 2D-Attention mechanism in~\cite{li2019show}. Both the Sequence Encoder and Decoder are 2-layer LSTM models with 512 hidden state size per layer.
At each encoding time step, the LSTM Encoder receives one column of $F(x)$ followed by max-pooling along the vertical axis, and updates its hidden state $h_{e}(x,t)$.
The final hidden state of the LSTM Encoder, $h_{e}(x,W)$, is provided for the LSTM Decoder as the initial state.
The attention mask is calculated by:
\begin{equation}
M^{'}_{ij}(x,t) = tanh(\sum_{p,q \in N(i,j)}{W_{F}F_{pq}(x)} + W_{h}h(x, t)),
\end{equation}
\begin{equation}
M(x,t) = sotfmax(W_{M}M^{'}),
\end{equation}
where $W_{F}$, $W_{h}$ and $W_{M}$ are linear transformations to be learned, and $N(i,j)$ is the neighborhood around position $(i,j)$ (i.e., $i-1 \leq p \leq i + 1$, $j-1 \leq q \leq j + 1$ ), $1 \leq i \leq H $, $1 \leq j \leq W $.

\begin{table}[!htbp]
\centering
\caption{The detailed configuration of our Glyph Generator and Discriminator.
``$k \times k \, (de)conv$'' means the kernel size of a (de)convolutional layer is $k$.
``s'' stands for stride of the (de)convolutional layer.
“Out Size” is the size of output feature maps of a block or a (de)convolutional layer (height $\times$ width $\times$ output channels).
The layers whose names are in bold receive the skip connections from CNN features.
}
\label{tb:CNNConfig}
\scalebox{1.00}{
\begin{tabular}{|p{1cm}|p{2cm}|p{4.2cm}|}
\hline
  Layers     & Out Size                 &  Configuration        \\
\hline
\multicolumn{3}{|c|}{CNN Feature Extractor}\\
\hline
\multicolumn{3}{|c|}{See Section~\ref{Sec:CNNModel}}\\
\hline
\multicolumn{3}{|c|}{Glyph Generator}\\
\hline
  Layer1     &   $2 \times 2 \times 128$           & $2 \times 2 \,deconv$, s $2 \times 2$    \\
  \hline
  Layer2     &   $4 \times 4 \times 64$           & $3 \times 3 \,deconv$, s $2 \times 2$    \\
  \hline
  \textbf{Layer3}     &   $8 \times 8 \times 32$           & $3 \times 3 \,deconv$, s $2 \times 2$    \\
  \hline
  \textbf{Layer4}     &   $16 \times 16 \times 16$           & $3 \times 3 \,deconv$, s $2 \times 2$    \\
  \hline
  \textbf{Layer5}     &   $32 \times 32 \times 1$           & $3 \times 3 \,deconv$, s $2 \times 2$    \\
\hline
\multicolumn{3}{|c|}{Glyph Discriminator}\\
\hline
  Layer1     &   $16 \times 16 \times 16$           & $3 \times 3 \,conv$, s $2 \times 2$    \\
  \hline
  Layer2     &   $8 \times 8 \times 32$           & $3 \times 3 \,conv$, s $2 \times 2$    \\
  \hline
  Layer3     &   $4 \times 4 \times 64$           & $3 \times 3 \,conv$, s $2 \times 2$    \\
  \hline
  Layer4     &   $2 \times 2 \times 128$           & $3 \times 3 \,conv$, s $2 \times 2$    \\
  \hline
  Layer5     &   $1 \times 1 \times 128$           & $3 \times 3 \,conv$, s $2 \times 2$    \\
  \hline
  Layer6     &   $1 \times 1 \times 1$           & $1 \times 1 \,conv$, s $1 \times 1$    \\
\hline
\end{tabular}
}
\end{table}

\subsection{Multi-font Glyph Generation}
The Glyph Generator is composed of a bunch of deconvolution layers, mapping the glimpse vector into glyphs progressively.
The highlights of our proposed glyph generation method are two-folded: (1) font embeddings are optimized along with the network's parameters synchronously.
They better serve as the font styles of generated glyphs so that the CNN and attention module can concentrate on extracting font-independent features from scene texts.
(2) an attentional and hierarchical generation framework is introduced which is effective for generating high-quality glyph images. \par
\textbf{Trainable font embeddings.} As shown in Equation~\ref{equ:GlyGen}, the font style of generated glyphs is controlled by the font embedding $z_{k}$.
We want to make sure $F(x)$ has captured enough reliable content features of the input character, so as to let $z_{k}$ manipulate the font style of generated glyphs.
$z_{1},z_{2},...,z_{m}$ are treated as trainable variables and will be fine-tuned according to the font styles of selected glyphs.
We adopt gradient descent to optimize font embeddings, which will be discussed in Section~\ref{Sec:OptFontEmbed}.
Through optimizing the font embeddings, our model concretize various font styles into a meaningful feature space.\par
\begin{figure}[t!]
  \centering
  \includegraphics[width=8cm]{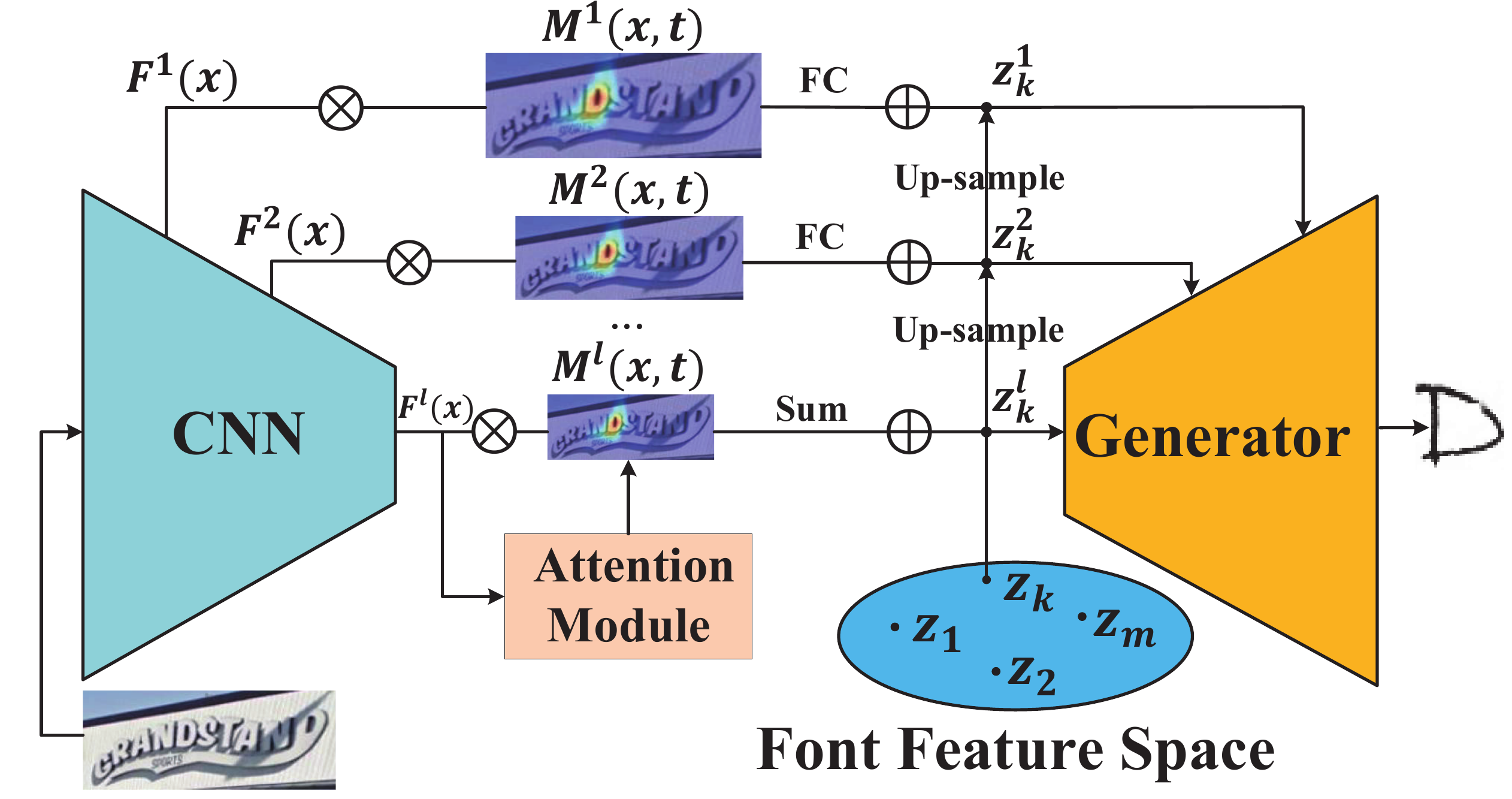}
  \caption{Our proposed attentional glyph generation.
  Multi-scale features from CNN are first multiplied by the attention masks, then concatenated with the font embeddings and finally sent into the Glyph Generator.
  $\bigotimes$ denotes point-wise multiplication over all channels and $\bigoplus$ denotes feature concatenation.}
  \label{fig:Generator}
\end{figure}

\textbf{Font-aware attentional skip connection.}
Multi-scale features from CNN are utilized for more accurately reconstructing the glyph's shape, which is shown in Figure~\ref{fig:Generator}.
Let $i$ ($1 \leq i \leq l$) be the index of selected multi-scale features from CNN.
The CNN features $F^{i}(x)$ are first multiplied by the attention mask $M^{i}(x,t)$, then transformed by fully connection or summation over all channels, afterwards concatenated with the font embedding $z^{i}_{k}$ , and finally sent into the Glyph Generator.
Note that $M^{l}(x,t) = M(x,t)$ and $z^{l}_{k} = z_{k}$.
$M^{i}(x,t)$ and  $z^{i}_{k}$ ($1 \leq i \leq l -1$) are up-sampled from the original $M(x,t)$ and $z_{k}$, respectively.
For the attention masks, the up-sampling operation is implemented by bilinear interpolation into the shape of $H_{i}\times W_{i}$, where $H_{i}$ and $W_{i}$ are the height and width of $F^{i}(x)$, respectively.
For the font embeddings $z^{i}_{k}$, the up-sampling operation is implemented by duplicating $z_{k}$ by $2^{l-i} \times 2^{l-i}$ times.
The “FC” operation (for $1 \leq i \leq l-1$) transforms the feature maps of $F^{i}(x) \bigotimes M^{i}(x,t)$ into the shape of $2^{l-i}\times 2^{l-i}$ by full connection.
The “Sum” operation (for $i=l$) means the feature map of each channel is reduced to $1 \times 1$ by summation.
\par
\textbf{Glyph Discriminator.}
Following the architecture of GANs, a Glyph Discriminator is introduced to distinguish between generated glyphs and real glyphs.
The Glyph Discriminator is a lightweight CNN followed by a fully connected layer with sigmoid activation.
Given a real (ground-truth) glyph or a generated glyph as input, it outputs a single value interpreted as the probability of the input glyph being real:
$p(y_{d} = 1 | g_{k}(x,t)\, or\, \hat{g}_{k}(x,t))$.
The Glyph Generator tries to maximize $p(y_{d} = 1 | \hat{g}_{k}(x,t))$ while the Glyph Discriminator tries to minimize $p(y_{d} = 1 | \hat{g}_{k}(x,t))$ and maximize $p(y_{d} = 1 | g_{k}(x,t))$.
Through the adversarial game between them, the quality of generated glyphs can be continuously improved.
The detailed configuration of the Glyph Generator and Glyph Discriminator can be found in Table~\ref{tb:CNNConfig}.
\par

\subsection{Loss Functions}
The basic loss function is composed of two items (the cross-entropy loss and the L1 loss):
\begin{equation}
L = - \sum_{t=1}^{T}\log p(y_{t} | x) +   \sum_{t=1}^{T} \Vert \hat{g}_{i_{t}}(x,t) - g_{i_{t}}(x,t)\Vert,
\end{equation}
where $y_{1}$, ...,  $y_{i}$, ..., $y_{T}$ are the ground-truth character labels represented in image $x$;
$i_{t}$ is a random integer sampled from $\{1,2,...,m\}$;
$ g_{i_{t}}(x,t)$ is the glyph in font $i_{t}$ of the $t$-th character in image $x$.
We find that this sampling method for glyph generation not only reduces the computation cost but also helps to achieve good performance.\par
When implementing adversarial training, there are two objective functions to be optimized iteratively:
\begin{equation}
\begin{split}
L_{G} = &- \sum_{t=1}^{T}[\log p(y_{t} | x) + \alpha \log p(y_{d} = 1 | \hat{g}_{i_{t}}(x,t))] \\
&+ \sum_{t=1}^{T}\Vert \hat{g}_{i_{t}}(x,t) - g_{i_{t}}(x,t)\Vert,
\end{split}
\end{equation}
\begin{equation}
L_{D} = - \alpha \sum_{t=1}^{T} [\log p(y_{d} = 0 | \hat{g}_{i_{t}}(x,t)) +  \log p(y_{d} = 1 | g_{i_{t}}(x,t))],
\end{equation}
where $\alpha$ is a hyper-parameter and set as $0.01$.
\subsection{Optimizing Font Embeddings}
\label{Sec:OptFontEmbed}
We optimize font embeddings with gradient descent, following the equation:
\begin{equation}
\begin{split}
& \frac{\partial L}{\partial z_{k}} = \sum_{t=1}^{T} \mathds{1} \{i_{t} = k\} \frac{\partial \Vert\hat{g}_{i_{t}} - g_{i_{t}}\Vert}{\partial z_{i_{t}}} \\
& = \sum_{t=1}^{T} \mathds{1}\{i_{t} = k\} \frac{\partial \Vert f(W_{c}c+W_{z}z_{i_{t}}) - g_{i_{t}}\Vert}{\partial z_{i_{t}}} \\
&= \sum_{t=1}^{T} \mathds{1} \{i_{t} = k\} ( W_{z} f^{'}(W_{c}c+W_{z}z_{i_{t}}) sgn(\hat{g}_{i_{t}} - g_{i_{t}})  ),
\end{split}
\end{equation}
where $\hat{g}_{i_{t}}$, ${g}_{i_{t}}$, and $c$ are the abbreviations of  $\hat{g}_{i_{t}}(x,t)$, ${g}_{i_{t}}(x,t)$, and $c(x,t)$, respectively;
$\mathds{1}$, $sgn$ and $f$ are the indicator function, sign function and activation function, respectively;
$W_{c} $ and $W_{z}$ are the parameters of Glyph Generator which are applied to $c$ and $z_{i_{t}}$, respectively;
$f^{'}$ is the derivative of $f$.
In practice, the Glyph Generator is composed of a bunch of deconvolutional layers and takes multi-scale features from CNN as input.
Here we formulate $\hat{g}_{i_{t}}$ as $f(W_{c}c+W_{z}z_{i_{t}})$ for brevity.
$W_{c}$ and $W_{z}$ are also trainable variables and hence the optimization of $W_{c}, W_{z}$ and $z_{1},...,z_{m}$ are alternate.
The computation of $ \frac{\partial L_{G}}{\partial z_{k}}$ is in the same way.
\section{Experiments}
We conduct extensive experiments to verify the effectiveness of our model and compare its performance with other state-of-the-art methods.
\begin{table*}[!htbp]
\centering
\caption{Recognition accuracy (in percentages) on public benchmarks in lexicon-free mode.
“90k”, “ST” and “SA” denote Synth90k, SynthText and SynAdd datasets, respectively. “ST$^{*}$” denotes that the character location information is exploited in SynthText. “R” denotes the training datasets of IC13, IC15 and COCO-T.
The best performing result for each dataset is shown in bold. Our approach achieves the best recognition performance on most benchmark datasets.
\cite{shi2018aster} have revised their result on SVT from 93.5 to 89.5 (see https://github.com/bgshih/aster).}
\label{tb:RecognitionAccuracy}
\scalebox{1.00}{
\begin{tabular}{|p{3.6cm}|p{2.5cm}|p{1cm}|p{1cm}|p{1cm}|p{1cm}|p{1cm}|p{1cm}|p{1.4cm}|}
\hline
  Method                                              & Training Data            &  IIIT5k         & SVT     & IC13  & IC15 & SVTP  & CT80  & COCO-T    \\
\hline
  Jaderberg et al.~\cite{jaderberg2015deep}    &90k                 &  -             & 71.7    &81.8    &  -    &  -  &  -  &  -        \\
\hline
  Jaderberg et al.~\cite{jaderberg2016reading}&90k                &  -          & 80.7         &90.8    &  -    &  -  &  -   &  -        \\
\hline
  Shi et al.~\cite{shi2017an} : CRNN       &90k               &  81.2              & 82.7    &89.6    &  -    &  -  &  - &  -         \\
\hline
  Lee et al.~\cite{lee2016recursive} : R2AM       &90k               &  78.4          & 80.7   &90.0    &   -    &  -   &  -   &  -     \\
\hline
  Liu et al.~\cite{liu2018synthetically}      &90k       &  89.4          &  87.1   & 94.0      &  -      &  -    &  -   &  -  \\
\hline
  Cheng et al.~\cite{cheng2017focusing} : FAN       &90k+ST$^{*}$    &  87.4       &  85.9  & 93.3  & 70.6   &  71.5   &  63.9    &  -     \\
\hline
  Liu et al.~\cite{liu2018char} : Char-Net              &90k+ST          &  92.0              &  85.5  & 91.1  & 74.2   &  78.9       &  -  &  -       \\
\hline
  Bai et al.~\cite{bai2018edit} : EP              &90k+ST          &  88.3              &  87.5  & 94.4  & 73.9   & -        &  -   &  -      \\
\hline
  Cheng et al.~\cite{cheng2018aon} : AON           & 90k+ST          &  87.0         &  82.8  & -  & 68.2   &  73.0   &  76.8   &  -       \\
\hline
  Shi et al.~\cite{shi2018aster} : ASTER            & 90k+ST          &  93.4            & 89.5$^{*} $   & 91.8   & 76.1  & 78.5 &  79.5 &  -         \\
\hline
   Zhan et al.~\cite{Zhan_2019_CVPR} : ESIR        & 90k+ST         &  93.3  &  90.2   & 91.3    & 76.9   & 79.6  &    83.3  &  -\\
\hline
   Wang et al.~\cite{wang2019boosting}         & 90k+ST         &  94.0  &  -   & 94.4    & -   & -  &    - &  -\\
\hline
  Ours                                 & 90k+ST      &  94.4  &  89.8   &  93.7   & 75.1   &  80.2  &  86.8  &  -\\
\hline
  Li et al.~\cite{li2019show} : SAR                & 90k+ST+SA+R     &  95.0        &  91.2  & 94.0  & 78.8  & \textbf{86.4}    & \textbf{89.6}       &  66.8           \\
\hline
  Ours                                 & 90k+ST+SA+R    &  \textbf{95.8}  &  \textbf{91.3}   & \textbf{95.1}    & \textbf{80.9}   & 86.0   &    88.5     &  \textbf{68.4}         \\
\hline
\end{tabular}
}
\end{table*}

\subsection{Datasets}
There exist two publicly available synthetic datasets that are widely used to train text recognizers: \textbf{Syn90k} released by~\cite{jaderberg2016reading} and \textbf{SynthText} proposed by~\cite{gupta2016synthetic}.
To compensate the lack of spatial characters in Syn90k and SynthText, ~\cite{li2019show} synthesized additional 1.6 million word images (denoted as \textbf{SynthAdd}).\par
The real-world datasets include \textbf{IIIT5K}~\cite{mishra2012top}, \textbf{SVT}~\cite{wang2011end}, \textbf{IC13}~\cite{karatzas2013icdar}, \textbf{IC15}~\cite{karatzas2015icdar}, \textbf{SVTP}~\cite{phan2013recognizing}, \textbf{CT80}~\cite{risnumawan2014a} and \textbf{COCO-T}~\cite{veit2016coco}.
The testing images of these datasets are benchmarks for evaluating the performance of STR models.
\par
\begin{figure}[!h]
  \centering
  \includegraphics[width=6cm]{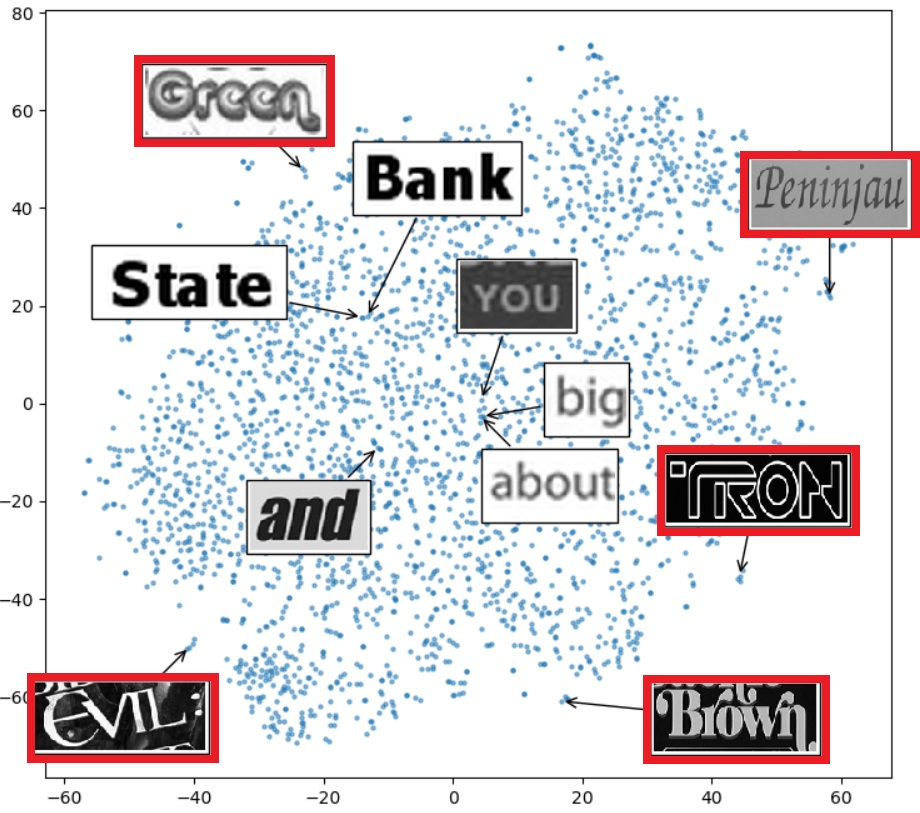}
  \caption{The distribution map of font features of text images in the IIIT5k dataset.}\label{fig:fontdistribution}
\end{figure}
\begin{figure}[t!]
  \centering
  \includegraphics[width=\columnwidth]{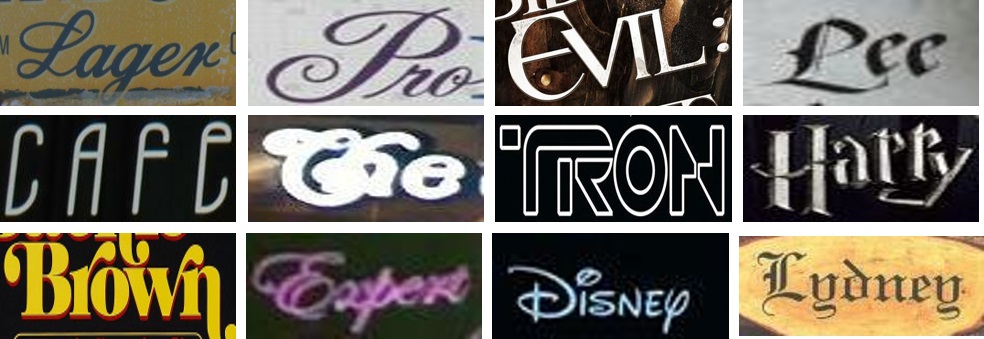}
  \caption{Some samples in our NFST (Novel Font Scene Text) dataset. The text images in NFST possess unusual and novel font styles and are difficult to be recognized by existing methods.}
  \label{fig:NFSTdataset}
\end{figure}

As scene texts in novel font styles only make up a small proportion in above-mentioned testing datasets, our method's superiority shown in Table~~\ref{tb:RecognitionAccuracy} is not that significant.
From IIIT5k , IC13, IC15, SVTP and COCO-T datasets , we collect 100 text images with novel or unusual font styles to form a new dataset named as the Novel Font Scene Text (\textbf{NFST}) dataset (see Figure~\ref{fig:NFSTdataset}).
Here we give a detailed description of how we select images for NFST.
Firstly, we consider a novel font as an eccentric and unusual font.
A distribution map of font features of candidate images is estimated to help our selection, which is shown in Figure~\ref{fig:fontdistribution}.
Specifically, we employ a font recognizer~\cite{wang2015deepfont} to extract the font features of all candidate text images.
The extracted features are then reduced into two dimensions via t-SNE.
Afterwards, these text images are attached according to the coordinates of their reduced features.
We select the most eccentric cases in the figure, where some examples are marked in red rectangles.
Specifically, we mainly select text images which have considerably less neighbors in the figure than others.
Meanwhile, we avoid selecting text images which are blurry, distorted, very small, etc.
In this manner, we believe that our NFST can be served as a good benchmark to measure the font-robustness of STR models.\par
To fully explore font-independent features, we should select those most representative fonts as targets, in terms of their popularity and style diversity, for our model to learn.
In our experiments, we choose 325 fonts ($m = 325$) from the \textbf{Microsoft Typography}\footnote{The font list can be found in https://docs.microsoft.com/en-us/typography/font-list} library.
Character samples `A' rendered by these fonts are shown in Figure~\ref{fig:FontEmbeddingsPCA}.
In our experiments, the image of each glyph is rendered with the font size 64 and resized into the resolution of $32 \times 32$.
\subsection{Implementation Details}
The proposed model is implemented in Tensorflow and trained on two NVIDIA 1080ti GPUs in parallel.
The whole network is end-to-end trained using the ADAM optimizer~\cite{kingma2015adam}.
The learning rate is set to $10^{-3}$ initially, with a decay rate of $0.9$ every 20000 iterations until it reaches $10^{-5}$.
The input images are resized to the resolution of 48 $\times$ 160.
The dimension of font embeddings $C^{'}$ is set as 128.
\subsection{Comparison with Other Methods}
\begin{figure}[t!]
  \centering
  \includegraphics[width=\columnwidth]{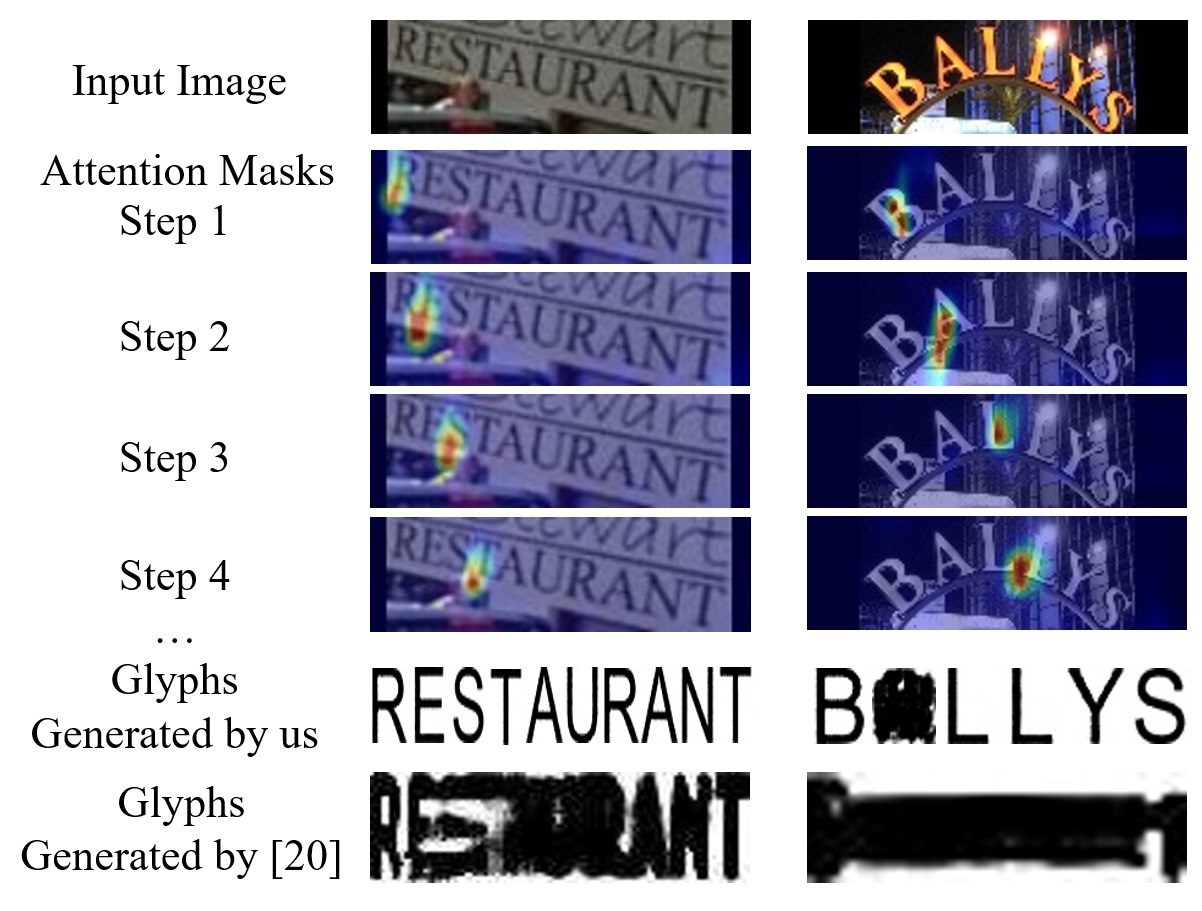}
  \caption{The spatial attention mechanism effectively assists our model with the glyph generation for irregular scene texts.
  Attention heat maps (visualization of the weights in attention masks) in the first 4 decoding steps are presented in this figure.
  In contrast, the method proposed by~\cite{liu2018synthetically} typically fails in coping with them.
  The glyph images generated by~\cite{liu2018synthetically} are quoted from their paper.
  The glyph images (in Arial font) generated by our model are re-scaled and arranged horizontally for a better view.}
  \label{fig:CompareGen}
\end{figure}

\begin{figure}[t!]
  \centering
  \includegraphics[width=\columnwidth]{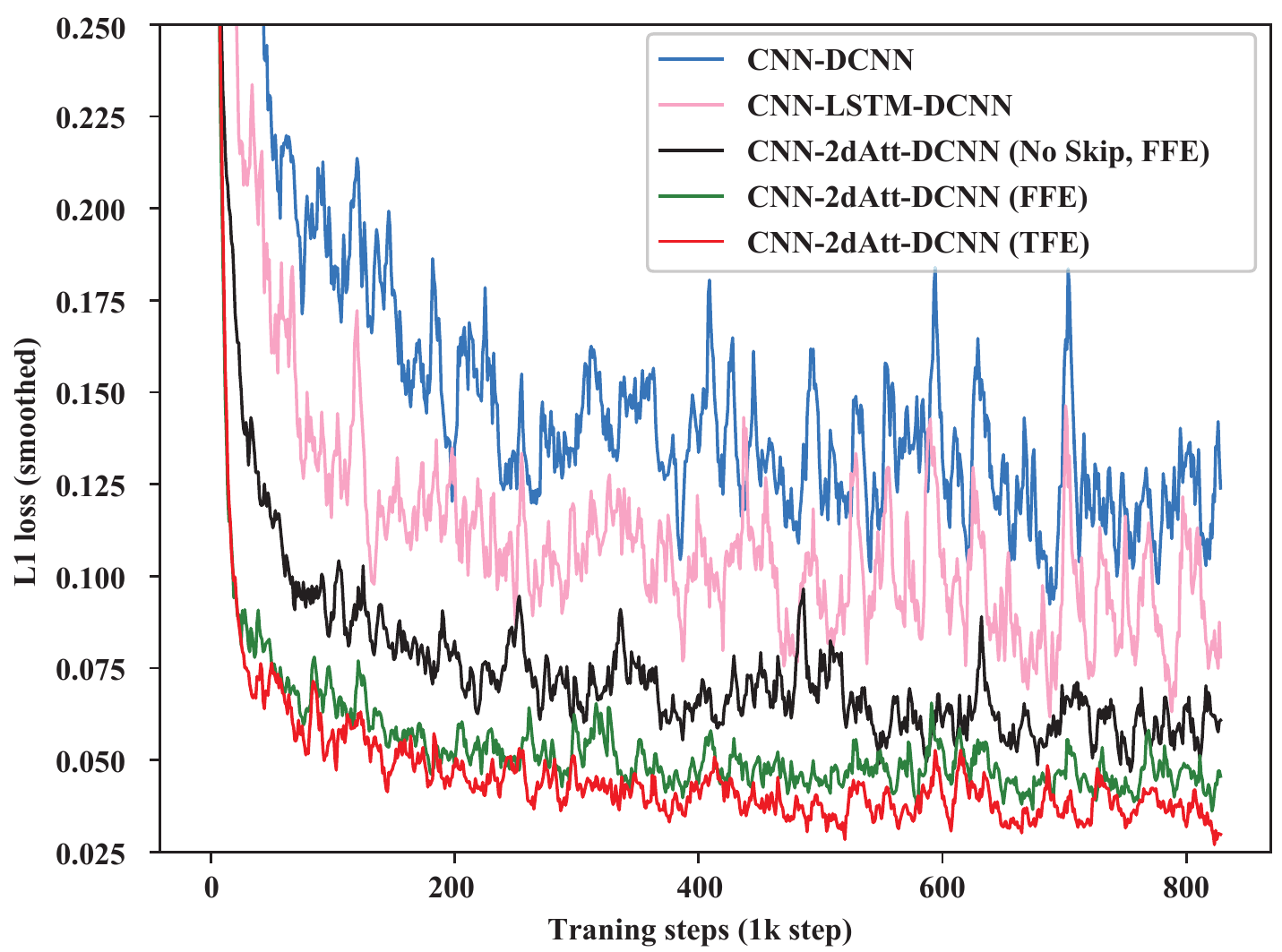}
  \caption{The curves in that L1 losses of different frameworks vary with training steps.
  Compared to other frameworks that are commonly seen in image-to-image translation, our proposed framework reconstructs more accurate glyphs of scene texts.}
  \label{fig:CompareL1Losses}
\end{figure}

\begin{figure}[t!]
  \centering
  \includegraphics[width=\columnwidth]{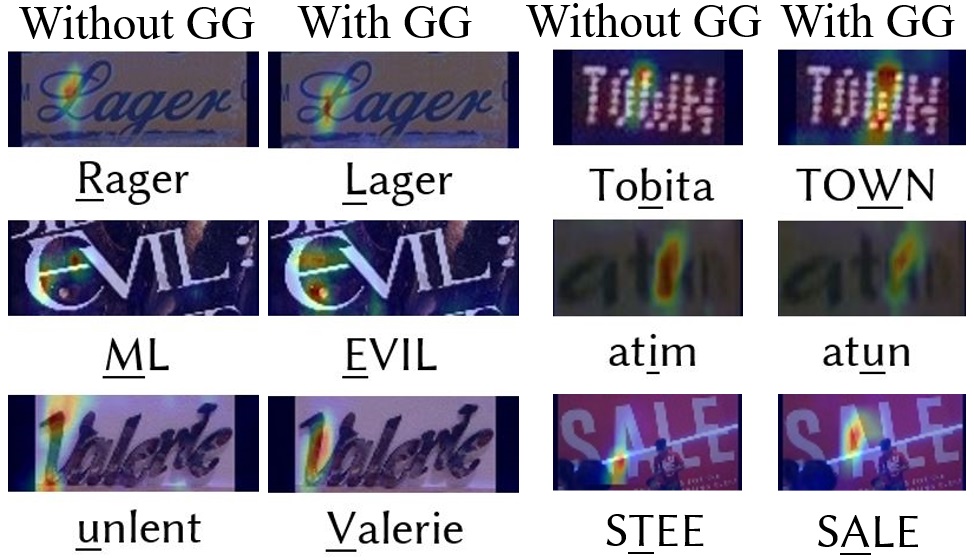}
  \caption{The proposed model shows more accurate perception on the discriminative parts of single character, in addition to the location of different characters. ``GG'' denotes Glyph Generation. The heat map and underlined character show the weight of attention mask and predicted character label of a certain step. }
  \label{fig:LocalShape}
\end{figure}
\textbf{Higher-quality generated glyphs.}
We compare the averaged L1 losses of five different generative frameworks, including CNN-DCNN, CNN-LSTM-DCNN, CNN-2DAtt-DCNN (No Skip, FFE), CNN-2DAtt-DCNN (FFE) and  CNN-2DAtt-DCNN (TFE).
The CNN-DCNN framework is employed in~\cite{liu2018synthetically}.
The CNN-LSTM-DCNN framework is an improved version of CNN-DCNN, where $F(x)$ is first sent into another LSTM encoder and then into DCNN, which is employed in~\cite{wang2019boosting}.
CNN-2DAtt-DCNN is our proposed attentional glyph generation framework.
``No Skip'' denotes no skip connection, i.e., only the CNN's last-layer features are utilized to generate glyphs.
FFE and TFE denote fixed font embeddings and trainable font embeddings, respectively.
For fair comparison, these frameworks are built with the same CNN, DCNN and LSTM configurations proposed in this paper and fed with the same training data (90k+ST+SA+R and glyphs in 325 fonts).
Fixed font embeddings are also deployed in CNN-DCNN and CNN-LSTM-DCNN.
The curves in the training stage are shown in Figure~\ref{fig:CompareL1Losses}, in which our method significantly outperforms others (lower is better).
This demonstrates that our model generates more accurate glyphs under the guidance of trainable font embeddings and 2D attention mechanism.
Our model can effectively cope with irregular texts which cannot be handled by~\cite{liu2018synthetically} (see Figure~\ref{fig:CompareGen}).
The generated glyphs directly reflect the quality of extracted CNN features:
for the right image in Figure~\ref{fig:CompareGen}, our model correctly recognizes it as ``BALLYS'' while ~\cite{liu2018synthetically} recognizes it as ``setes''.\par
\textbf{More precise perception on local shapes.}
Accurate location on the discriminative parts of each single character is also very important for correct recognition.
Figure~\ref{fig:LocalShape} shows that our model achieves a better perception on them after being trained to generate glyphs.
Without glyph generation (GG), our model is a 2D attention based text recognizer, sharing similar architecture with~\cite{li2019show}.
Taking the first case for example, without the guidance of GG, our model pays its attention to the upper part of `L' in round hand (see the red positions in heat map), which wrongly recognize it as `R'.
After introducing the generation of multi-font glyphs, our model focuses on the lower part of `L' and correctly recognize it.
The key lies in that our model can successfully exclude undesired nuisance factors from semantic features.\par
\textbf{Higher recognition accuracy.}
Owing to the positive impacts of the proposed approach, our model achieves the best performance on most benchmark datasets (shown in Table~\ref{tb:RecognitionAccuracy}).
On SVTP and CT80, which are relatively small datasets containing 639 and 288 images respectively, our model performs slightly worse than~\cite{li2019show}.
Generally, our model achieves the state-of-the-art performance compared to other models.
Although we adopt the spatial attention module proposed in~\cite{li2019show}, our proposed generative learning branch further improves the recognition performance.
Our method remarkably improves the recognition accuracy on the IC15 dataset, where many scene texts are placed in cluttered environments and possess novel font or writing styles.\par
\textbf{Recognizing scene texts in novel styles.}
In this section we conduct a quantitative experiment on the NFST dataset to demonstrate our model’s robustness to the variance of font styles.
We compare our method with other two state-of-the-art methods~\cite{shi2018aster,li2019show} whose codes are publicly available.
Our method significantly outperforms others on this dataset (see Table~\ref{tb:NFSTRecognition}), whose robustness to font style variance is proved.\par

\begin{table}[!htbp]
\centering
\caption{Recognition accuracy (in percentages) of different methods on the NFST dataset.}
\label{tb:NFSTRecognition}
\begin{tabular}{|p{2.5cm}|p{1cm}|p{1.2cm}|p{1.7cm}|}
\hline
  Training data        &  Ours         & SAR~\cite{li2019show}     & ASTER~\cite{shi2018aster}     \\
\hline
  90k+ST            & \textbf{55.0}         &   45.0      &  44.0              \\
\hline
  90k+ST+SA+R      & \textbf{71.0}         &    63.0         &  58.0              \\
\hline
\end{tabular}
\end{table}
\subsection{The Optimization Process of Font Embeddings}
\begin{figure}[t!]
  \centering
  \includegraphics[width=\columnwidth]{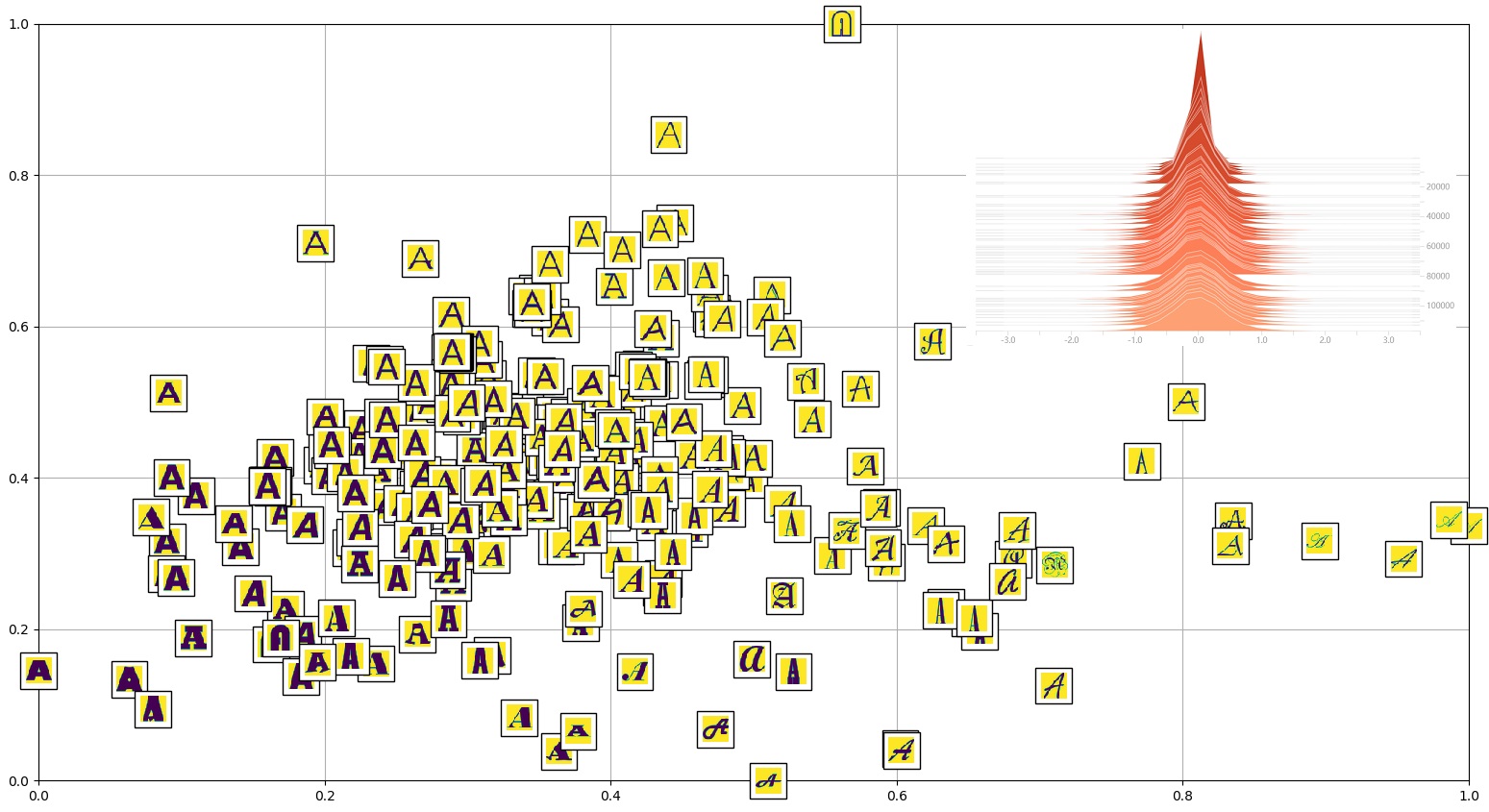}
  \caption{Visualization of all trained font embeddings that are reduced into two dimensions with PCA. The glyphs `A' in different fonts are attached according to the coordinates of corresponding font embeddings.
  In the top-right corner, we show how the distribution histogram of values in font embeddings varies with training steps (the vertical axis).}
  \label{fig:FontEmbeddingsPCA}
\end{figure}

In this section, we illustrate the optimization process of font embeddings to have a better understanding of how they work.
As mentioned in the previous sections, the font embeddings are in charge of controlling the font styles of glyphs.
Thereby, the font embeddings need to represent the actual distribution of font styles of selected glyphs.
In the top-right corner of Figure~\ref{fig:FontEmbeddingsPCA}, we show how the distribution of values in font embeddings varies with the training steps.
The font embeddings are randomly initialized, obeying the Gaussian distribution $N(0,0.01)$.
It changes significantly in the early stage to fit the actual distribution of all font styles.
Afterwards, it remains relatively stable in the training phase.
Figure~\ref{fig:FontEmbeddingsPCA} also demonstrates the visualization of all trained font embeddings that are reduced into two dimensions with PCA (Principal Component Analysis).
The glyphs `A' in different fonts are attached according to the coordinates of corresponding font embeddings.
We can observe that similar fonts are located closer while dissimilar fonts are located farther away.
Those novel fonts tend to locate at the border of the 2D distribution space.
This figure clearly demonstrates that our font embeddings are endowed with actual meanings through training.
\par

\begin{figure}[t!]
  \centering
  \includegraphics[width=\columnwidth]{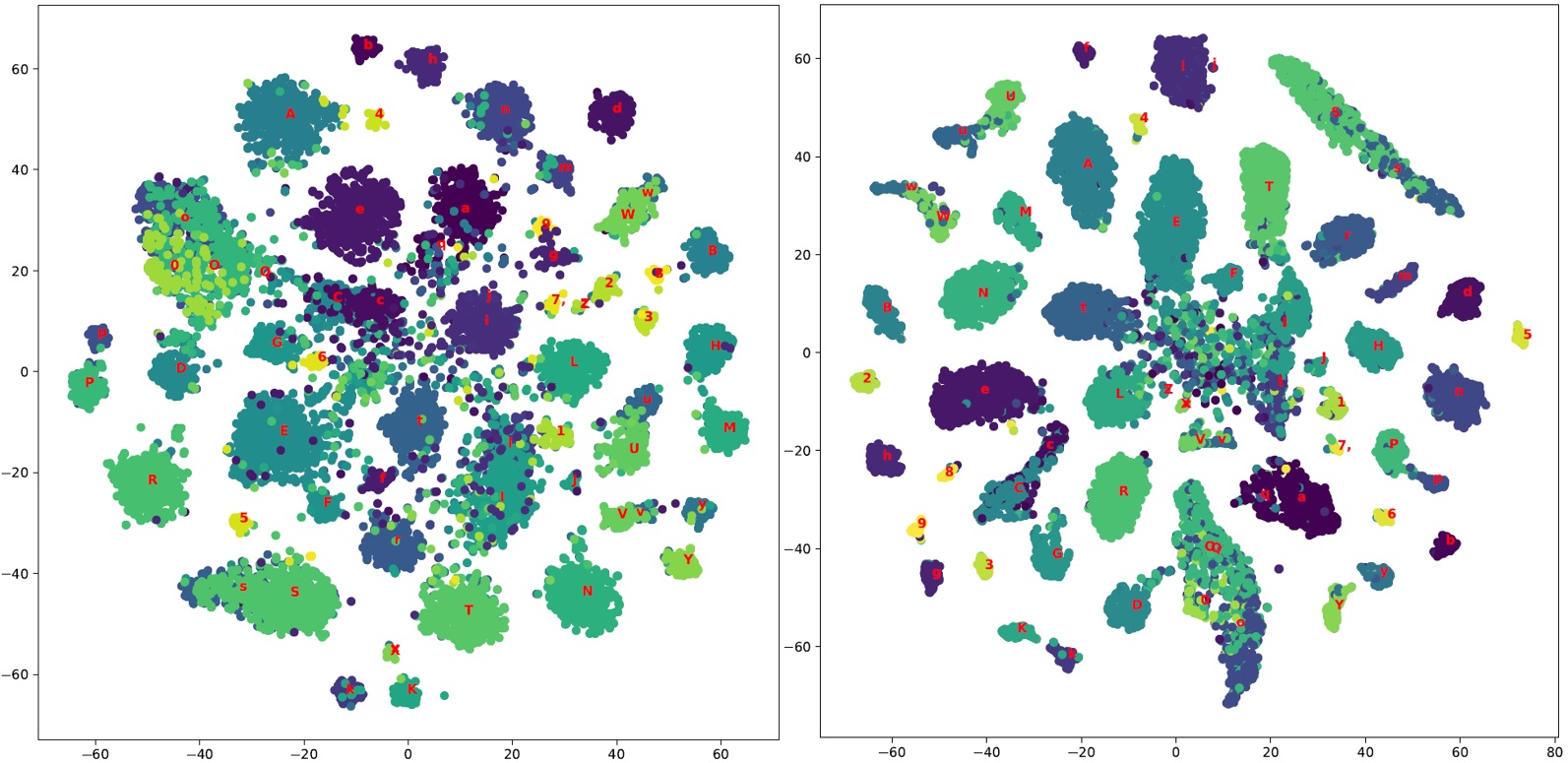}
  \caption{The visualization result of CNN's last-layer features of our model without (left) and with (right) glyph generation. The experiment is conducted on the IIIT5K testing dataset. After introducing glyph generation, the features of different characters with similar shapes are less tangled, such as `9' and `g'.
  The feature clusters of different characters are more clearly separated and the ambiguous cases are less scattered.}
  \label{fig:FeatTSNEVis}
\end{figure}

\subsection{Ablation Study}
\label{Sec:AblaStudy}
\begin{figure}[t!]
  \centering
  \includegraphics[width=\columnwidth]{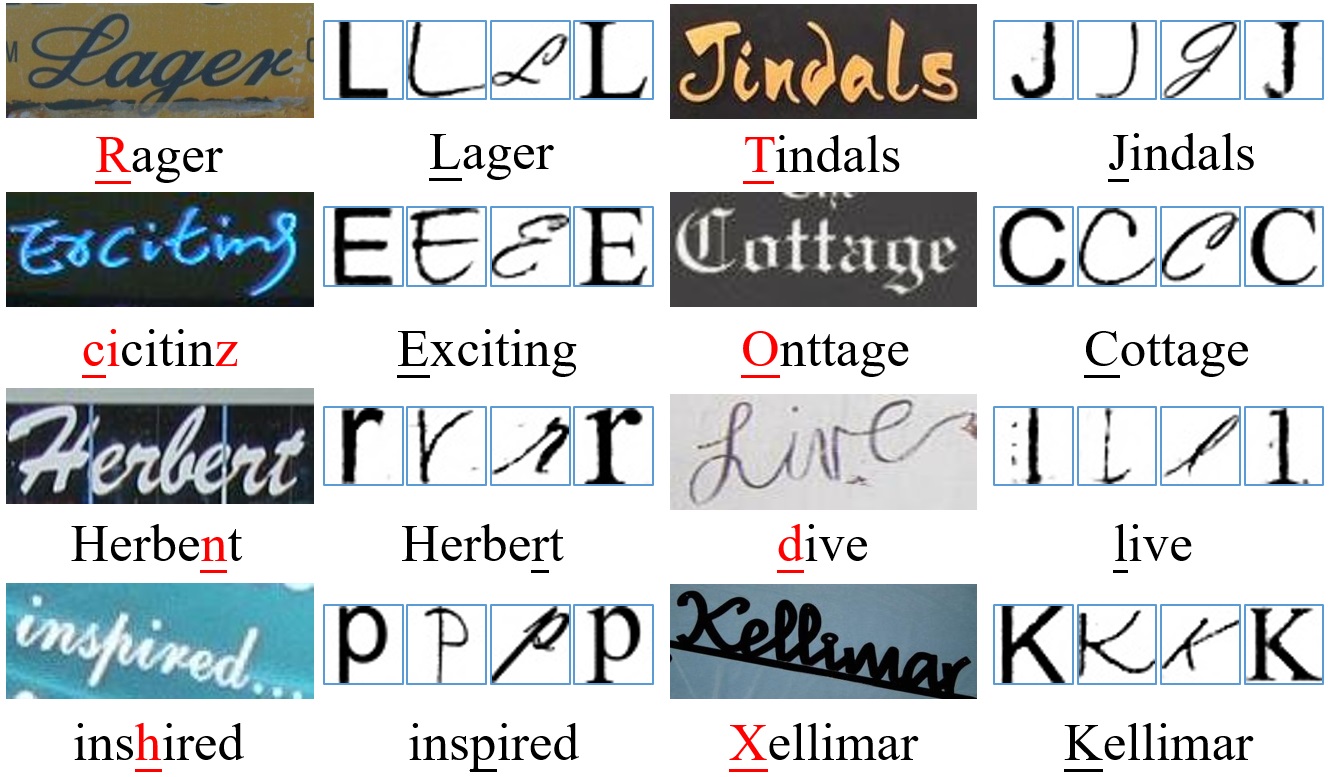}
  \caption{Text samples in the NFST dataset which were wrongly recognized with fixed font embeddings but correctly recognized with trainable font embeddings.
  The red underline characters show the wrong predictions and next to them our predictions and generated glyphs are demonstrated.}
  \label{fig:EffectofTFE}
\end{figure}

For the purpose of analyzing the impacts of different modules on the recognition performance, we conduct a series of ablation studies as shown in Table~\ref{tb:AblationStudy}.
``Att-1 and Att-2'' denotes our model which employs the attention mechanism employed in~\cite{shi2018aster} and~\cite{sheng2018nrtr} respectively.
``-GD, -GG'' denotes our model in which Glyph Generator and Glyph Discriminator are both removed (i.e., no glyph generation).
``FFE'' denotes our model which employs fixed font embeddings.
Without glyph generation, our model's architecture is similar to~\cite{li2019show} thus the performance is also nearly the same.
After introducing glyph generation, the learnt features of different kinds of characters become less tangled, which is demonstrated in Figure~\ref{fig:FeatTSNEVis} by utilizing t-SNE~\cite{maaten2008visualizing}.
In Figure~\ref{fig:FeatTSNEVis}, each cluster consists of features of character images in different fonts but the same character category.
The clusters are more centralized and outliers are significantly reduced in our model, which justifies the claim of font-independent features.
Our model understands characters' font styles more deeply and performs better by introducing trainable font embeddings (more results are shown in Figure~\ref{fig:EffectofTFE}).
To investigate how our model's performance varies with the number of fonts ($m$), we train our model with different settings of $m$ ($10, 50, 100, 325$).
For $m = {10, 50, 100}$, we employ the K-means algorithm for the learnt 325 embeddings to find $m$ representative fonts as the new targets for our model to generate.
\begin{table}[!htbp]
\centering
\caption{Results of ablation studies by removing or changing the proposed modules in our model.}
\label{tb:AblationStudy}
\begin{tabular}{|p{1.5cm}|p{0.8cm}|p{0.8cm}|p{0.8cm}|p{0.8cm}|p{0.8cm}|}
\hline
  Method        &  IIIT5k         & SVT     & IC13  & IC15 & NFST  \\
\hline
  Att-1   &  95.4    & 90.9 & 94.7   & 79.8 &  -  \\
\hline
  Att-2   &  95.5  & 91.2    & 94.9   & 80.5 &  -  \\
\hline
  -GG, -GD & 95.0       &  91.1   & 94.1   & 78.5   &  - \\
\hline
  FFE     &  95.3        &  91.1   & 94.5   &   80.3  &  61.0 \\
\hline
  10fonts          &  95.5        & 91.1    & 94.8   &  80.1 &  59.0  \\
\hline
  50fonts          &  95.5        & 91.1    & 94.8   &  80.3 &  63.0  \\
\hline
  100fonts          &  95.6        & 91.2    & 94.9   &  80.7 & 67.0  \\
\hline
  full          &  95.8        & 91.3    & 95.1   &  80.9 & 71.0  \\
\hline
\end{tabular}
\end{table}

\section{Conclusion}
In this paper, we proposed the attentional glyph generation with trainable font embeddings for improving the feature learning of scene text recognition.
Trainable font embeddings make significant contributions for removing the font information from the CNN features of scene texts.
Besides, we proposed to utilize the spatial attention mechanism for glyph generation, which effectively deals with irregular scene texts.
Experimental results demonstrated that our model generates higher-quality glyphs, acquires more precise perception on text shapes and achieves higher recognition accuracy compared to the state of the art.
In the future, we will try to apply more advanced GAN-related techniques on the Glyph Generator and Discriminator to further improve the quality of generated glyphs.
There is also much room for improving the spatial attention mechanism to better locate each character in text images.

\begin{acks}
This work was supported by National Natural Science Foundation of China (Grant No.: 61672056 and 61672043), Beijing Nova Program of Science and Technology (Grant No.: Z191100001119077), Center For Chinese Font Design and Research, and Key Laboratory of Science, Technology and Standard in Press Industry (Key Laboratory of Intelligent Press Media Technology).
\end{acks}

\bibliographystyle{ACM-Reference-Format}
\bibliography{myreference}

\end{document}